\documentclass[11pt,a4paper]{article}
\usepackage[hyperref]{emnlp-ijcnlp-2019}
\usepackage{times}
\usepackage{latexsym}

\usepackage{url}
\usepackage{amsfonts} 
\usepackage{booktabs} 
\usepackage{multirow} 
\usepackage{xspace}
\usepackage{graphicx}
\usepackage{color}
\usepackage{colortbl}
\usepackage[english]{babel}
\usepackage{balance}
\usepackage{amsmath,bm}
\usepackage{makecell}
\usepackage{algorithm}
\usepackage{algpseudocode}
\usepackage{amssymb}
\usepackage{subcaption}

\definecolor{Gray}{gray}{0.9}

\newcommand\cparagraph[1]{\vspace{1mm}\noindent\textbf{#1.}}

\definecolor{Gray}{gray}{0.9}

\aclfinalcopy 

\title{Jointly Learning Entity and Relation Representations \\for Entity Alignment}

\author{
	Yuting Wu$^{1}$, 
	Xiao Liu$^{1}$, 
	Yansong Feng$^{1,2}$\thanks{\;\;Corresponding author.},  
	Zheng Wang$^{3}$ \and
	Dongyan Zhao$^{1,2}$ \\
	$^1$Wangxuan Institute of Computer Technology, Peking University, China\\
	$^2$The MOE Key Laboratory of Computational Linguistics, Peking University, China\\
	$^3$School of Computing, University of Leeds, U.K. \\
	{\tt \{wyting,lxlisa,fengyansong,zhaodongyan\}@pku.edu.cn} \\
	{\tt z.wang5@leeds.ac.uk}\\
}

\date{}

\begin{document}
\maketitle

	\begin{abstract} Entity alignment is a viable means for integrating heterogeneous knowledge among different knowledge graphs (KGs). Recent developments in the field often take an embedding-based approach to model the structural information of KGs so that entity alignment can be easily performed in the embedding space. However, most existing works do not explicitly utilize useful relation representations to assist in entity alignment, which, as we will show in the paper, is a simple yet effective way for improving entity alignment. This paper presents a novel joint learning framework for entity alignment. At the core of our approach is a Graph Convolutional Network (GCN) based framework for learning both entity and relation representations. Rather than relying on pre-aligned relation seeds to learn relation representations, we first approximate them using entity embeddings learned by the GCN. We then incorporate the relation approximation into entities to iteratively learn better representations for both. Experiments performed on three real-world cross-lingual datasets show that our approach substantially outperforms state-of-the-art entity alignment methods.
	
\end{abstract}

\section{Introduction}
\label{section:intro}
Knowledge graphs (KGs) transform unstructured knowledge into simple and clear triples of $<$\emph{head entity, relation, tail entity}$>$
for rapid response and reasoning of knowledge. They are an effective way for supporting various NLP-enabled tasks like machine reading \cite{yang2017kblstm}, information extraction \cite{wang2018label}, and question-answering
\cite{zhang2018variational}.

Even though many KGs originate from the same resource, e.g., Wikipedia, they are usually created independently. Therefore, different KGs often use different expressions and surface forms to
indicate equivalent entities and relations -- let alone
those built from different resources or languages. This common problem of heterogeneity makes it difficult to integrate knowledge among different KGs. 
A powerful technique to address this issue is \emph{Entity Alignment}, the task of linking entities with the same real-world identity from
different KGs.

Classical methods for entity alignment typically involve a labor-intensive and time-consuming process of feature construction
\cite{mahdisoltani2013yago3} or rely on external information constructed by others \cite{Suchanek:2011:PPA:2078331.2078332}. Recently, efforts have been devoted to
the so-called embedding-based approaches. Representative works of this direction include JE \cite{hao2016joint}, MTransE
\cite{chen2016multilingual}, JAPE \cite{sun2017cross}, IPTransE \cite{zhu2017iterative}, and BootEA \cite{ijcai2018-611}. More recent work
\cite{D18-1032} uses the Graph Convolutional Network (GCN) \cite{Kipf2016Semi} to jointly embed multiple KGs.

Most of the recent works (e.g., JE, MTransE, JAPE, IPTransE and BootEA) rely on the translation-based models, such as TransE
\cite{bordes2013translating}, 
which enable these approaches to encode both entities and relations of KGs. These methods often put more emphasis on the entity embeddings, but do not explicitly utilize relation embeddings to help with entity alignment. Another
drawback of such approaches is that they usually rely on pre-aligned relations (JAPE and IPTransE) or triples (MTransE). This limits the
scale at which the model can be effectively performed due to the overhead for constructing seed alignments for large KGs. Alternative
methods like GCN-based models, unfortunately, cannot directly obtain relation representations, leaving much room for improvement.

Recent studies have shown that jointly modeling entities and relations in a single framework can improve tasks like information extraction
\cite{miwa-bansal-2016-end,bekoulis-etal-2018-adversarial}. We hypothesize that this will be the case for entity alignment too; that is,
the rich relation information could be useful for improving entity alignment as entities and their relations are usually
closely related. Our experiments show that this is even a conservative target: by jointly learning entity and relation representations,
we can promote the results of both entity and relation alignment.

In this work, we aim to build a learning framework that jointly learns entity and relation representations for entity alignment; and we
want to achieve this with only a small set of pre-aligned entities but not relations. Doing so will allow us to utilize relation information
to improve entity alignment without paying extra cost for constructing seed relation alignments.

Our work is enabled by the recent breakthrough effectiveness of GCNs \cite{Kipf2016Semi} in extracting useful representations from graph
structures. Although GCNs provide a good starting point, applying it to develop a practical and efficient framework to accurately capture
relation information across KGs is not trivial. Because a vanilla GCN operates on the undirected and unlabeled graphs, a GCN-based model
like \cite{D18-1032} would ignore the useful relation information of KGs. While the Relational Graph Convolutional Network (R-GCN)
\cite{Schlichtkrull2017Modeling} can model multi-relational graphs, existing R-GCNs use a weight matrix for each relation. This means that an R-GCN would require an excessive set of parameters to model thousands of relations in a typical real-world KG, making it difficult
to learn an effective model on large KGs.

A key challenge of our joint learning framework is how to generate useful relation representations at the absence of seed relation
alignments, and to ensure the framework can scale to a large number of types of relations. We achieve this by first approximating the
relation representations using entity embeddings learned through a small amount of seed entity alignments. We go further by constructing a
new joint entity representation consisting of both relation information and neighboring structural information of an entity. The joint
representations allow us to iteratively improve the model's capability of generating better entity and relation representations, which lead
to not only better entity alignment, but also more accurate relation alignment as a by-product.

We evaluate our approach by applying it to three real-world datasets. Experimental results show that our approach delivers better and more
robust results when compared with state-of-the-art methods for entity and relation alignments.
The key contribution of this paper is a novel joint learning model for entity and relation alignments. Our approach reduces the human
involvement and the associated cost in constructing seed alignments, but yields better performance over prior works.

\section{Related Work}
\subsection{Entity Alignment\label{sec:kgalignment}}

Until recently, entity alignment would require intensive human participation \cite{VrandecicKroetzsch14cacm} to design hand-crafted
features \cite{mahdisoltani2013yago3}, rules, or rely on external sources \cite{Wang2017}. In a broader context, works in schema and
ontology matching also seek help from additional information by using e.g., extra data sources~\cite{Nguyen:2011:MSM:2078324.2078329}, entity
descriptions~\cite{lacoste2013sigma,yang2015entity}, or semantics of the web ontology language~\cite{hu2011self}. Performance of such
schemes is bounded by the quality and availability of the extra information about the target KG, but obtaining sufficiently good-quality
annotated data could be difficult for large KGs.

Recently, embedding-based entity alignment methods were proposed to reduce human involvement. JE \cite{hao2016joint} was among the first attempts in this
direction. It learns embeddings of different KGs in a uniform vector space where entity alignment can be performed. MTransE
\cite{chen2016multilingual} encodes KGs in independent embeddings and learns transformation between KGs. BootEA \cite{ijcai2018-611}
exploits a bootstrapping process to learn KG embeddings. SEA \cite{Pei:2019:SEA:3308558.3313646} proposes a degree-aware KG embedding model to embed KGs. KDCoE \cite{Chen2018Co} is a semi-supervised learning approach for co-training embeddings for multilingual KGs and entity descriptions. They all use translation-based models as the backbone to embed KGs. 

Non-translational embedding-based methods include recent works on a GCN-based model \cite{D18-1032}
and NTAM \cite{ijcai2018-581}. Additionally, most recent work, RDGCN \cite{ijcai2019-733}, introduces the dual relation graph to model the relation information of KGs. Through multiple rounds of interactions between the primal and dual graphs, RDGCN can effectively incorporate more complex relation information into entity representations and achieve promising results for entity alignment. However, existing methods only focus on entity embeddings and ignore the help that relation representations can provide on this task.

MTransE and NTAM are two of a few methods that try to perform both relation and entity alignments. However,
both approaches require high-quality seed alignments, such as pre-aligned triples or relations, for relation alignment.
Our approach advances prior works by jointly modeling entities and relations by using only a \emph{small set} of pre-aligned entities (\emph{but not relations}) to simultaneously perform entity and relation alignments.

\subsection{Graph Convolutional Networks} 
GCNs \cite{Duvenaud2015Convolutional,Kearnes2016Molecular,Kipf2016Semi} are neural networks operating on unlabeled graphs and inducing
features of nodes based on the structures of their neighborhoods. Recently, GCNs have demonstrated promising performance in tasks like
node classification \cite{Kipf2016Semi}, relation extraction \cite{zhang2018graph}, semantic role labeling \cite{Marcheggiani2017Encoding}, etc. As an extension of GCNs, the R-GCNs \cite{Schlichtkrull2017Modeling} have recently been proposed to model relational data for link
prediction and entity classification. However, R-GCNs
usually require a large number of parameters that are often hard to train, when applied to multi-relational graphs.

In this work, we choose to use GCNs to first encode KG entities and to approximate relation representations based on entity embeddings. Our
work is the first to utilize GCNs for jointly aligning entities and relations for heterogeneous KGs.

\section{Problem Formulation\label{sec:problem}}
We now introduce the notations used in this paper and define the scope of this work.

A KG is formalized as $G = (E, R, T)$, where $E, R, T$ are the sets of entities, relations and triples, respectively. Let $G_1 = (E_1, R_1,
T_1)$ and $G_2 = (E_2, R_2, T_2)$ be two different KGs. Usually, some equivalent entities between KGs are already known,
defined as \emph{alignment seeds} $\mathbb{L} = \{(e_{i_1}, e_{i_2}) | e_{i_1} \in E_1, e_{i_2} \in E_2\}$.

We define the task of entity or relation alignment as automatically finding more equivalent entities or relations based on known alignment
seeds. In our model, we only use known \emph{aligned entity pairs} as training data for both entity and relation alignments. The process of
relation alignment in our framework is unsupervised, which does not need pre-aligned relation pairs for training.

\section{Our Approach}\label{section:app}

\begin{figure*}[t!]
	\centering
	\includegraphics[width=1.0\textwidth]{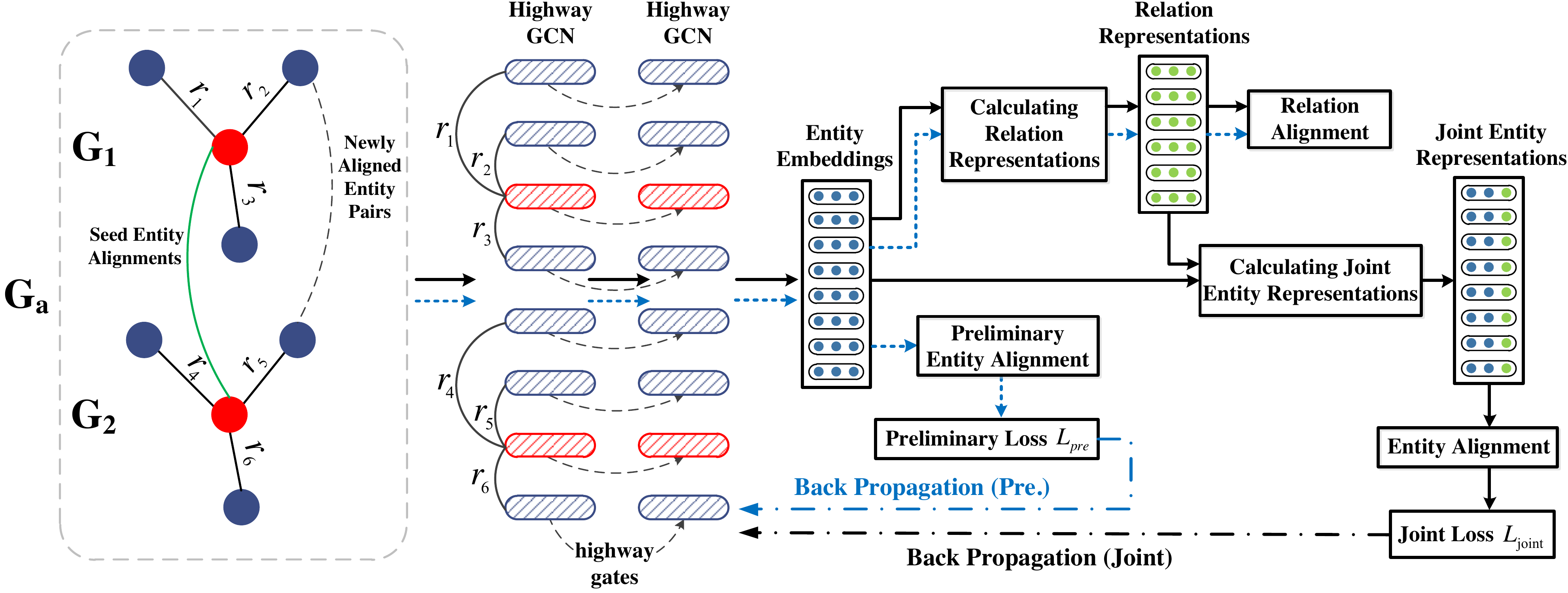}
	\caption{Overall architecture of our model. The blue dotted lines denote the process of preliminary entity alignment and
		preliminary relation alignment using approximate relation representations, and the black solid lines denote the process of continuing using GCNs to iteratively learn better entity and relation representations.}\label{fig:all}
\end{figure*}

Given two target KGs, $G_1$ and $G_2$, and a set of known aligned entity pairs $\mathbb{L}$, our approach uses GCNs \cite{Kipf2016Semi}
with highway network \cite{Srivastava2015Highway} gates to embed entities of the two KGs and approximate relation semantics based on
entity representations. By linking entity representations with relation representations, they promote each other in our framework and ultimately achieve better alignment results.

\subsection{Overall Architecture}
\label{section:overall}

As illustrated in Figure \ref{fig:all}, our approach consists of three stages: (1) preliminary entity alignment, (2) approximating relation representations, and (3) joint entity and relation alignment.

In the first stage, we utilize GCNs to embed entities of various KGs in a unified vector space for preliminary entity alignment. Next, we use the entity embeddings to
approximate relation representations which can be used to align relations across KGs. In the third stage, we incorporate the relation representations into entity embeddings to obtain the joint entity representations, and continue using GCNs to iteratively integrate neighboring
structural information to achieve better entity and relation representations.

\subsection{Preliminary Entity Alignment}
\label{preEntAlign}
As shown in Figure \ref{fig:all}, we put $G_1$ and $G_2$ in one graph $G_a=(E_a, R_a, T_a)$ to form our model's input. We utilize pre-aligned
entity pairs to train our model and then discover latent aligned entities.

\cparagraph{Graph Convolutional Layers} Our entity alignment model utilizes GCNs to embed entities in $G_a$. Our model consists of multiple
stacked GCN layers so that it can incorporate higher degree neighborhoods. The input for GCN layer $l$ is a node feature matrix, $\textbf{X}^{(l)}
=\{\bm{x}^{(l)}_1,\bm{x}^{(l)}_2,...,\bm{x}^{(l)}_{n} |\bm{x}^{(l)}_{i} \in \mathbb{R}^{d^{(l)}}\}$, where $n$ is the number of nodes (entities) of $G_a$, and
$d^{(l)}$ is the number of features in layer $l$. $\textbf{X}^{(l)}$ is updated using forward propagation as:
\begin{equation}
\textbf{X}^{(l+1)} = \mathrm{ReLU}(\tilde{D}^{- \frac{1}{2}}\tilde{A}\tilde{D}^{- \frac{1}{2}}\textbf{X}^{(l)}\textbf{W}^{(l)}),
\end{equation}
where $\tilde{A}=A+I$ is the adjacency matrix of $G_a$ with self-connections, $I$ is an identity matrix,
$\tilde{D}_{jj}=\sum_k\tilde{A}_{jk}$, and $\textbf{W}^{(l)} \in \mathbb{R}^{d^{(l)}
	\times d^{(l+1)}}$ is a layer-specific trainable weight matrix.

Inspired by \cite{Rahimi2018Semi} that uses highway gates \cite{Srivastava2015Highway} to control the noise propagation in GCNs for geographic localization, we also employ layer-wise highway gates to build a \emph{Highway}-GCN
(HGCN) model. Our layer-wise gates work as follow:
\begin{equation}
T(\textbf{X}^{(l)})=\sigma(\textbf{X}^{(l)}\textbf{W}_T^{(l)}+\bm{b}_T^{(l)}),\\
\end{equation}
\begin{equation}
\textbf{X}^{(l+1)}= T(\textbf{X}^{(l)}) \cdot \textbf{X}^{(l+1)}+(\textbf{1}-T(\textbf{X}^{(l)})) \cdot \textbf{X}^{(l)}
\end{equation}
where $\textbf{X}^{(l)}$ is the input to layer $l+1$; $\sigma$ is a sigmoid function; $\cdot$ is element-wise multiplication; $\textbf{W}_T^{(l)}$ 
and $\bm{b}_T^{(l)}$ 
are the weight matrix and bias vector for the transform gate $T(\textbf{X}^{(l)})$,
respectively.

\cparagraph{Alignment\label{prediction}} In our work, entity alignment is performed by simply measuring the distance between two entity nodes on their embedding space.
With the output entity representations
$\textbf{X}' = \{\bm{x}'_1,\bm{x}'_2,...,\bm{x}'_n|\bm{x}'_{i} \in \mathbb{R}^{\tilde{d}}\}$, for entities $e_1$ from $G_1$ and $e_2$ from $G_2$, their distance is
calculated as:
\begin{equation}
\label{d}
d(e_1,e_2)=\|\bm{x}'_{e_1}-\bm{x}'_{e_2}\|_{L_1}.
\end{equation}

\cparagraph{Training} 
We use a margin-based scoring function as the training objective, to make the distance between aligned entity pairs to be as close as possible, and the distance between positive and negative alignment pairs to be as large as possible. The loss function is defined as:
\begin{equation}
\label{trainingobj}
L=\sum\limits_{(p,q)\in \mathbb{L}}\sum\limits_{(p',q')\in \mathbb{L'}}\mathrm{max}\{0,d(p,q)-d(p',q')+\gamma\},
\end{equation}
where $\gamma>0$ is a margin hyper-parameter; $\mathbb{L'}$ stands for the negative alignment set of $\mathbb{L}$.

Rather than simply random sampling for negative instances, we look for more challenging negative samples, e.g., those with subtle differences from the positive ones, to train our model.
Given a positive aligned pair $(p,q)$,  
we choose the $\mathcal{K}$-nearest
entities of $p$ (or $q$) according to Eq. \ref{d} in the embedding space to replace $q$ (or $p$) as the negative instances. 

\subsection{Approximating Relation Representations}
\label{subsection:relationAlign}
At this stage, we expect to obtain relation representations, which can be used in the next stage for constructing joint representations and can also be used for preliminary relation alignment.
Since we are unable to explicitly modeling relations within our GCN-based framework, we thus approximate the relation representations based
on their head and tail entity representations produced by the entity alignment model described in Section~\ref{preEntAlign}. This strategy is
based on our observation that the statistical information of the head and tail entities of a relation can more or less reflect the shallow
semantics of the relation itself, such as the head or tail entities' type requirements of a relation. Our experiments in Section~\ref{sec:results} suggest that
this is a reasonable assumption.

Given a relation $r \in R_a$, there is a set of triples of $r$, $\mathbb{T}_r = \{(h_i, r, t_j) | h_i \in H_r, t_j \in T_r\}$, where $H_r$ and $T_r$ are the sets of head entities and tail entities of relation $r$, respectively. For a relation $r$, its representation can be approximated as:
\begin{equation}
\label{relEmbeddings}
\bm{r} = f(\textbf{H}_r, \textbf{T}_r),
\end{equation}
where $\bm{r}$ is the approximated representation of relation $r$. $\textbf{H}_r$ and $\textbf{T}_r$ are the sets of HGCN-output embeddings of head
entities and tail entities of relation $r$. $f(\cdot)$ is a function to produce relation representations with input entity vectors, which can
take many forms such as \emph{mean}, \emph{adding}, \emph{concatenation} or more complex models. In our model, we compute
the relation representation for $r$ by first concatenating its averaged head and tail entity representations, and then introducing a matrix
$\textbf{W}_R \in \mathbb{R}^{2\tilde{d} \times m}$ as a learnable shared linear transformation on relation vectors. Here, $\tilde{d}$ is the number
of features in each HGCN-output entity embedding and $m$ is the number of features in each relation representation.

With the relation representations in place, relation alignment can be performed by measuring the distance between two relation vectors. For relation $r_1$ from $G_1$ and $r_2$ from $G_2$, their distance is computed as:
\begin{equation}
\label{rel_distance}
s(r_1, r_2) = \|\bm{r}_1-\bm{r}_2\|_{L_1}-\beta\frac{|P_{r_1r_2}|}{|HT_{r_1} \cup HT_{r_2}|},
\end{equation}
where $\bm{r}_1$ and $\bm{r}_2$ are the relation representations for $r_1$ and $r_2$. In addition to calculating the distance between the two relation vectors, we believe that the more aligned entities exist in the entities that are connected to the two relations, the more likely the two relations are equivalent. Thus, for $r_1$ and $r_2$, we collect the pre-aligned entities existing in the head/tail entities of these two relations as the set $P_{r_1r_2} = \{(e_{i_1}, e_{i_2}) | e_{i_1} \in HT_{r_1}, e_{i_2} \in HT_{r_2}, (e_{i_1}, e_{i_2}) \in \mathbb{L}\}$. $HT_{r_1}$ and $HT_{r_2}$ are the sets of head/tail entities for relation $r_1$ and $r_2$ respectively. $\beta$ is a hyper-parameter for balance.

In our framework, relation alignment is explored in an unsupervised fashion, in which we do not have any pre-aligned relations as training data.

\subsection{Joint Entity and Relation Alignment}
\label{subsection:joint}
The first two stages of our approach could already produce a set of entity and relation alignments, but we do not stop here.
Instead, we attentively fuse the entity and relation representations and further jointly optimize them using the seed entity alignments.
Our key insight is that entity and relation alignment tasks are
inherently closely related. This is because aligned entities tend to have some relations in common, and similar relations should have
similar categories of head and tail entities.

Specifically, we first pre-train the entity alignment model (Section \ref{preEntAlign}) until its entity alignment performance has converged to be
stable. We assume that both the pre-trained entity and approximate relation representations can provide rich information for themselves.
Next, for each entity, we aggregate the representations of its relevant relations into
a relation context vector, which is further combined
with its pre-trained entity representation to form a new joint entity representation.

Formally, for each entity $e \in E_a$, its new joint representation  $\bm{e}_{joint}$ can be calculated as:
\begin{equation}
\label{jointEmbeddings}
\bm{e}_{joint} = g(\bm{e}, \textbf{R}_e),
\end{equation}
where $\bm{e}$ is the HGCN-output representation of entity $e$.
$\textbf{R}_e$ is the set of relation representations of $e$'s relevant relations. $g(\cdot)$ is a function to produce the new joint entity representation by taking $\bm{e}$ and $\textbf{R}_e$ as input, which can also take many forms of operations. In our model, we calculate
$\bm{e}_{joint}$ by first summing all relation representations in $\textbf{R}_e$ and then concatenating $\bm{e}$ with the summed relation context vector.

After getting the new joint entity representations, $\textbf{X}_{joint}$, we continue optimizing our model against the seed entity alignments, where we use the joint entity representations to calculate the training loss according to Eq. \ref{trainingobj} to continue updating HGCNs\footnote{The training procedure is detailed in Appendix A.}. Note that the joint entity representations are composed of entity embeddings and relation representations, while the relation representations are also constructed based on the entity embeddings. Hence, after backpropagation of the loss calculated using the joint entity representations, we optimize the entity embeddings.

\section{Experimental Setup}
\subsection{Datasets}
\label{subsection:datasets}
We use DBP15K datasets from \cite{sun2017cross} to evaluate our approach. DBP15K contains three cross-lingual datasets that were built from the English version to Chinese, Japanese and French versions of DBpedia. Each contains data from two KGs in different languages and provides 15K pre-aligned entity pairs. Besides, each dataset also provides some pre-aligned relations. We manually aligned more relations from the three datasets and removed the ambiguously aligned relation pairs to construct the test sets for relation alignment. Table \ref{dataset} shows the statistics of the three datasets. We stress that our approach achieves entity and relation alignments simultaneously using only a small number of pre-aligned entities, and relation alignments are only used for testing. Following the previous works \cite{sun2017cross,D18-1032,ijcai2018-611}, we use 30\% of the pre-aligned entity pairs as training data and 70\% for testing. Our source code and datasets are freely available online\footnote{https://github.com/StephanieWyt/HGCN-JE-JR}.

\begin{table}[t!]
	\centering
	\scriptsize
	\begin{tabular}{l|l|ccc|cc}
		\Xhline{1pt}
		\multicolumn{2}{c|}{\multirow{2}{*}{\bf DBP15K}} & \multirow{2}{*}{\textbf{\#Ent.}} & \multirow{2}{*}{\textbf{\#Rel.}} & \multirow{2}{*}{\textbf{\#Rel tr.}} & \multicolumn{2}{c}{\textbf{Alignments}}\\
		\multicolumn{1}{c}{} & \multicolumn{1}{c|}{} & & & & \textbf{\#Ent.} & \textbf{\#Rel.} \\
		\hline
		\multirow{2}{*}{ZH-EN} & ZH & 66,469 & 2,830 & 153,929 & \multirow{2}{*}{1,5000} & \multirow{2}{*}{890}\\
		& EN & 98,125 & 2,317 & 237,674 \\
		\hline
		\multirow{2}{*}{JA-EN} & JA & 65,744 & 2,043 & 164,373 & \multirow{2}{*}{1,5000} & \multirow{2}{*}{529} \\
		& EN & 95,680 & 2,096 & 233,319\\
		\hline
		\multirow{2}{*}{FR-EN} & FR & 66,858 & 1,379 & 192,191 & \multirow{2}{*}{1,5000} & \multirow{2}{*}{212} \\
		& EN & 105,889 & 2,209 & 278,590\\
		\Xhline{1pt}
	\end{tabular}
	\caption{Summary of the DBP15K datasets.}
	\label{dataset}
\end{table}

\subsection{Implementation Details}
\label{implementation} We set $\gamma = 1$, $\beta = 20$, and learning rate to 0.001. We sample $\mathcal{K}=125$ negative pairs every 50
epochs.
We use entity names in different KGs for better model initialization. We translate non-English entity names to English via Google Translate, and the entity features are initialized with pre-trained English word vectors
\emph{glove.840B.300d}\footnote{http://nlp.stanford.edu/projects/glove/} in our model.
Note that Google Translate does not always give accurate translations for named entities. We inspected 100 English translations for Japanese and
Chinese entity names, and discovered that around 20\% of the translations are wrong. The errors are mainly attributed to the missing of
titles/modifications and wrong interpretations for person/location names. The inaccurate translation poses further challenges for our model.

\subsection{Competitive Approaches}

\cparagraph{Entity Alignment}
For entity alignment, we compare our approach against six embedding-based entity alignment methods discussed in
Section \ref{section:intro}: JE \cite{hao2016joint}, MTransE \cite{chen2016multilingual}, JAPE \cite{sun2017cross}\footnote{We note
	that \cite{sun2017cross} also provides analysis by considering the outputs of a machine translator and JAPE, and using a theoretically
	perfect \emph{oracle} predictor to correctly choose in between the results given by the machine translator and JAPE. As this only
	serves as an interesting up-bound analysis, but does not reflect the capability of JAPE (because it is impossible to build such a perfect predictor in the first place),
	we do not compare to this oracle implementation. },
IPTransE \cite{zhu2017iterative}, BootEA \cite{ijcai2018-611} and GCN \cite{D18-1032}. Among those, BootEA is the best-performing model on DBP15K.

\cparagraph{Relation Alignment}
For relation alignment, we compare our approach with the state-of-the-art BootEA (denoted by BootEA-\emph{R}), and MTransE (denoted by MTransE-\emph{R}).
Note that MTransE provides five implementation variants for its alignment model. To provide a fair comparison, we choose the one that does
not use pre-aligned relations but gives the best performance for a triple-wise alignment verification \cite{chen2016multilingual} - a
closely related task for relation alignment. Since BootEA and MTransE are translation-based models that encode both entities and
relations, relation alignment can be done by measuring the similarities between two relation representations. Furthermore, to evaluate the
effectiveness of our proposed relation approximation method, we also build BootEA-\emph{PR} and MTransE-\emph{PR} for relation alignment according
to Section \ref{subsection:relationAlign}.

\subsection{Evaluation Methodology}
\cparagraph{Model Variants}
To evaluate our design choices, we provide different implementation variants with the following denotations.
HGCN is our base GCN model with highway gates and entity name initialization. It has several variants, described as follows. HGCN-\emph{PE
}(Section \ref{preEntAlign}) and HGCN-\emph{PR} (Section \ref{subsection:relationAlign}) are our preliminary models for entity and
relation alignments, respectively. HGCN-\emph{JE} and HGCN-\emph{JR} are our complete models that use joint representations to further improve entity alignment and relation alignment (Section \ref{subsection:joint}). Finally, GCN-\emph{PE} and GCN-\emph{PR} are the preliminary GCN-based models for entity and relation alignments respectively, which use
entity name initialization \emph{but no highway gates}; GCN-\emph{JE} and GCN-\emph{JR} are the corresponding joint learning models; and GCN-\emph{JE-r} is the randomly initialized version of GCN-\emph{JE} without entity name initialization.

\cparagraph{Metrics} Like prior works \cite{sun2017cross,D18-1032,ijcai2018-611}, we use Hits@k as our evaluation metric. A Hits@k score is computed by measuring the proportion of correctly aligned entities ranked in the top $k$ list. Hence, we prefer higher Hits@k scores that indicate better performance.

\section{Experiment Results\label{sec:results}}
In this section, we first show that our complete model consistently outperforms all alternative methods across datasets,
metrics and alignment tasks. We then analyze the impact of prior alignment data size on model performance, showing that our approach
requires significantly less training data but achieves better performance over the best-performing prior method. Finally, we use a concrete
example to discuss how jointly learned entity and relation representations can be used to improve both entity and relation alignments.

\subsection{Entity Alignment\label{sec:result:ea}}
Table \ref{Entresults} reports the performance for entity alignment of all compared approaches. The top part of the table shows the
performance of prior approaches. By using a bootstrapping process to expand the training data, BootEA clearly outperforms all prior
methods. By capturing the rich neighboring structural information, GCN outperforms all other translation-based models on Hits@1, and over IPTransE, MTransE and
JE on Hits@10.

The bottom part of Table \ref{Entresults} shows how our proposed techniques, i.e., entity name initialization, joint embeddings and
layer-wise highway gates, can be used within a GCN framework to improve entity alignment. After initialized with the machine-translated
entity names, GCN-PE considerably improves GCN on all datasets. The improvement suggests that even rough translations of entity names (see
Section \ref{implementation}) can still provide important evidence for entity alignment and finally boost the performance. By employing layer-wise highway
gates, HGCN-PE further improves GCN-PE, giving a 34.31\% improvement on Hits@1 on \textit{DBP15K}$_{FR-EN}$, and also outperforms the strongest baseline BootEA. This substantial improvement indicates that highway gates can effectively control the propagation of noisy information. Our complete
framework HGCN-JE gives the best performance across all metrics and datasets. Comparing HGCN-JE with HGCN-PE and GCN-JE with GCN-PE (2.36\%
and 4.19\% improvements of Hits@1 on \textit{DBP15K}$_{ZH-EN}$ respectively), we observe that joining entity and relation alignments
improves the model performance. Even without entity name initialization, GCN-JE-r still has obvious advantages over JE, MTransE, JAPE,
IPTransE and GCN. The results reinforce our claim that merging the relation information into entities can produce better entity
representations. We stress that our proposed methods are not restricted to GCNs or HGCNs, but can be flexibly integrated with other KG
representation models as well.
\begin{table}[t!]
	\centering
	\scriptsize
	\begin{tabular}{p{38pt}|p{14pt}<{\centering}p{18pt}<{\centering}|p{14pt}<{\centering}p{18pt}<{\centering}|p{14pt}<{\centering}p{18pt}<{\centering}}
		\toprule
		\multirow{2}{*}{\bf Models} & \multicolumn{2}{c|}{\bf ZH-EN} & \multicolumn{2}{c|}{\bf JA-EN} & \multicolumn{2}{c}{ \bf FR-EN}  \\
		& \tiny \bf Hits@1 & \tiny \bf Hits@10 & \tiny \bf Hits@1 &\tiny \bf Hits@10 & \tiny \bf Hits@1 &\tiny \bf Hits@10\\
		\midrule
		JE & 21.27 & 42.77 & 18.92 & 39.97 & 15.38 & 38.84\\
		MTransE & 30.83 & 61.41 & 27.86 & 57.45 & 24.41 & 55.55\\
		JAPE & 41.18 & 74.46 & 36.25 & 68.50 & 32.39 & 66.68 \\
		IPTransE& 40.59 & 73.47 & 36.69 & 69.26 & 33.30 & 68.54 \\
		BootEA  & 62.94 & 84.75 & 62.23 & 85.39 & 65.30 & 87.44 \\
		GCN & 41.25 & 74.38 & 39.91 & 74.46 & 37.29 & 74.49  \\
		\midrule
		\bf GCN-JE-r& 45.92 & 72.91 & 46.62 & 74.57 & 48.27 & 77.82 \\
		\bf GCN-PE & 50.70 & 79.23 & 53.01 & 82.83 & 54.42 & 84.65 \\
		\bf GCN-JE & 54.89 & 81.30 & 56.35 & 83.73 & 58.30 & 86.34\\
		\bf HGCN-PE & 69.67 & 82.56 & 75.50 & 87.89 & 88.73 & 95.52 \\
		\bf HGCN-JE & \bf 72.03 & \bf 85.70 & \bf 76.62 & \bf 89.73 & \bf 89.16 & \bf 96.11 \\
		\bottomrule
	\end{tabular}
	\caption{Performance on entity alignment.}
	\label{Entresults}
\end{table}

\begin{table}
	\centering
	\scriptsize
	\begin{tabular}{p{38pt}|p{14pt}<{\centering}p{18pt}<{\centering}|p{14pt}<{\centering}p{18pt}<{\centering}|p{14pt}<{\centering}p{18pt}<{\centering}}
		\toprule
		\multirow{2}{*}{\bf Models} & \multicolumn{2}{c|}{\bf ZH-EN} & \multicolumn{2}{c|}{\bf JA-EN} & \multicolumn{2}{c}{ \bf FR-EN}  \\
		& \tiny \bf Hits@1 & \tiny \bf Hits@10 & \tiny \bf Hits@1 &\tiny \bf Hits@10 & \tiny \bf Hits@1 &\tiny \bf Hits@10\\
		\midrule
		MTransE-R& 3.03 & 8.88 & 2.65 & 10.21 & 3.30 & 14.62 \\
		MTransE-PR& 32.81 & 57.64 & 31.00 & 56.14 & 18.87 & 44.34 \\
		BootEA-R& 55.17 & 70.00 & 47.83 & 67.67 & 36.79 & 58.49 \\
		BootEA-PR & 45.28 & 85.37 & 41.40 & 79.77 & 30.19 & \bf 60.38 \\
		\midrule
		\bf GCN-PR & 66.18 & 82.81 & 60.87 & 81.47 & 38.21 & 52.83 \\
		\bf GCN-JR & 70.22 & 84.38 & 63.89 & 81.10 & 41.98 & 53.77 \\
		\bf HGCN-PR & 69.33 & 84.49 & 63.14 & 81.26 & 41.51 & 54.25 \\
		\bf HGCN-JR & \bf 70.34 & \bf 85.39 & \bf 65.03 & \bf 83.55 & \bf 42.45 & 56.60 \\
		\bottomrule
	\end{tabular}
	\caption{Performance on relation alignment.}
	\label{Relresults}
\end{table}

\begin{figure*}[t!]
	\centering
	\includegraphics[width=1.0\linewidth]{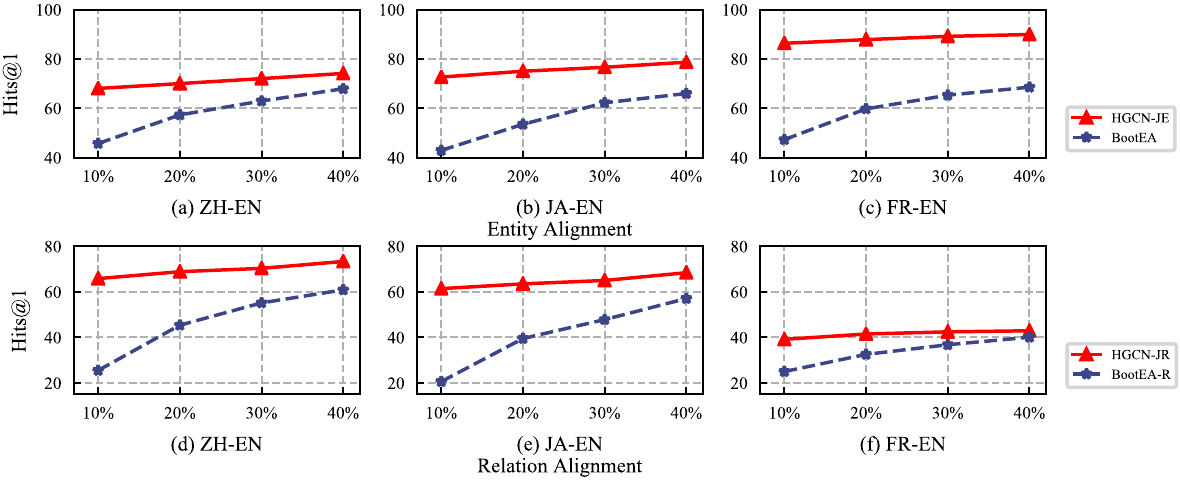}
	\caption{(a)-(c) report the performance for entity alignment of HGCN-JE and BootEA when they are trained with different proportions of seed entity alignments on the three DBP15K datasets. (d)-(f) show the relation alignment performance of HGCN-JR and BootEA-R under corresponding conditions. The x-axes are the proportions of seed alignments, and the y-axes are Hits@1 scores.}
	\label{six}
\end{figure*}

\subsection{Relation Alignment\label{sec:result:ra}}
Table \ref{Relresults} reports the results of relation alignment. Directly using the relation embeddings learned by MTransE to perform
relation alignment leads to rather poor performance for MTransE-R, less than 4\% for Hits@1 for all datasets. This is because the
translation assumption, $head+relation \approx tail$, used by MTransE focuses on modeling the overall relationship among heads, tails, and
relations, but capturing little neighboring information and relation semantics. After approximating the relation representations using
entity embeddings according to Eq \ref{relEmbeddings}, MTransE-PR substantially improves MTransE-R. This confirms our assumption that it is
feasible to approximate a relation using the information of its head and tail entities.

The strong entity alignment model BootEA also performs well for relation alignment.
Using the relation embeddings from BootEA, BootEA-R
delivers the best Hits@1 in MTransE and BootEA variants. Using our approximation strategy hurts BootEA-R in Hits@1, but we see improvements on Hits@10 across all datasets. This suggests that our approximation method can bring more related candidates, but may lack precision to select top-ranked candidates, comparing to explicitly relation modeling in translation-based models.

Our framework, HGCN-JR, delivers the best relation alignment results across datasets and metrics, except for Hits@10 on \textit{DBP15K}$_{FR-EN}$. Like entity alignment, we also observe
that joining entity and relation alignments improves relation alignment, as evidenced by the better performance of HGCN-JR and GCN-JR
over HGCN-PR and GCN-PR, respectively. That is, joint modeling produces better entity representations, which in turn provide better relation approximations. This can promote the results of both alignment tasks.

\subsection{Analysis\label{sec:prioralignmentdata}}
\cparagraph{Impact of Available Seed Alignments}
To explore the impact of the size of seed alignments on our model, we compare our HGCN with BootEA by varying the
proportion of pre-aligned entities from 10\% to 40\% with a step of 10\%. Figure \ref{six} (a-c) illustrate the Hits@1 for entity alignment of HGCN-JE and BootEA on three datasets. As the amount of seed alignments increases, the performances of both models on all three data sets gradually improve. HGCN-JE consistently obtains superior results compared to BootEA, and seems to be
insensitive to the proportion of seed alignments. For example, HGCN-JE still achieves 86.40\% for Hits@1 on \textit{DBP15K}$_{FR-EN}$ when only using 10\% of training data. This Hits@1 score is 17.84\% higher than that of BootEA when BootEA uses 40\% of seed alignments.

Figure \ref{six} (d-f) show the Hits@1 for relation alignment of HGCN-JR and BootEA-R. HGCN-JR also consistently outperforms BootEA-R, and gives more stable results with different ratios of seed entity alignments.
These results further confirm the robustness of our model, especially with limited seed entity alignments.

\begin{figure}[t!]
	\centering
	\includegraphics[width=1.0\linewidth]{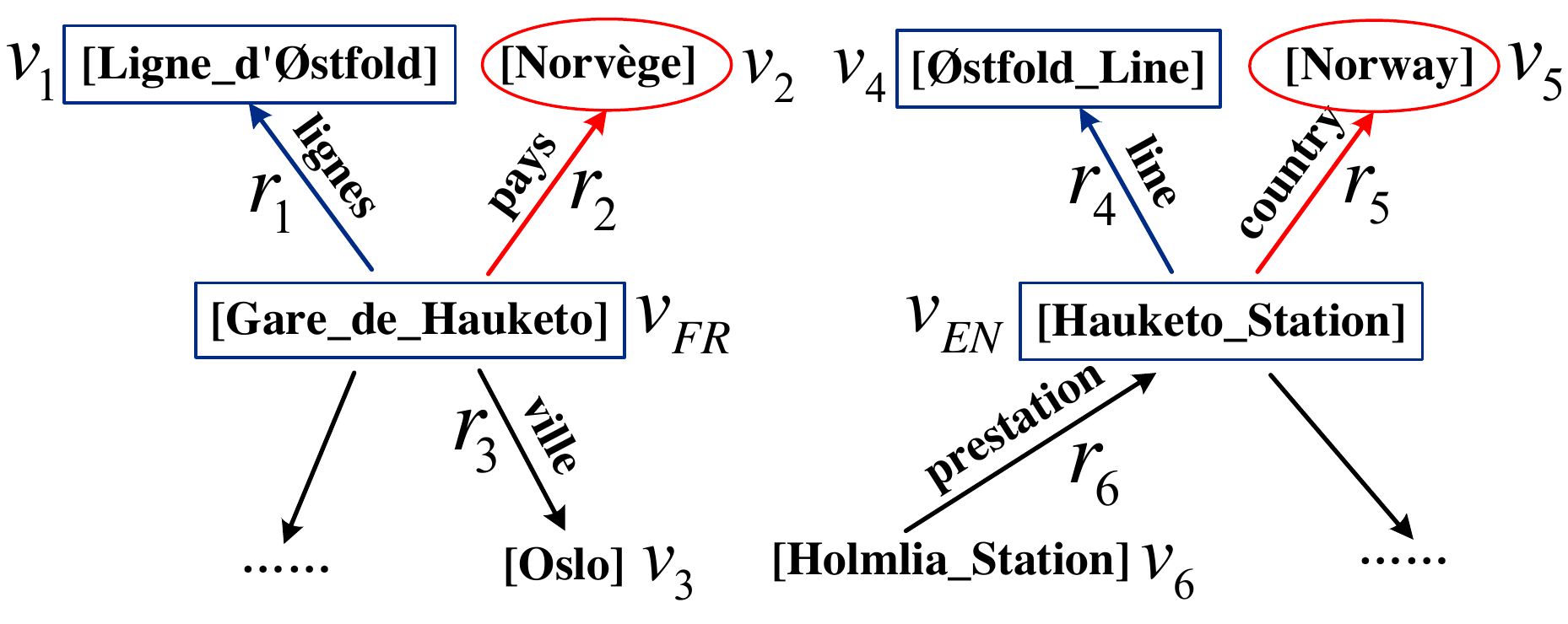}
	\caption{A real-world example from \textit{DBP15K}$_{FR-EN}$. $[v_2; v_5]$ and $[r_2$; $r_5]$ are respectively the aligned entities and aligned relations after performing preliminary entity and relation alignments. $[v_{FR}$; $v_{EN}]$ and $[v_1$; $v_4]$ are the newly aligned entity pairs, and $r_1$ and $r_4$ are the newly aligned relations, which are discovered using jointly learned entity and relation representations. Jointly optimizing alignment tasks leads to the sucessful discovery of new aligned relation and entity pairs.}
	\label{case}
\end{figure}

\cparagraph{Case Study}
Figure \ref{case} shows an example from \textit{DBP15K}$_{FR-EN}$. In the stages of preliminary entity alignment and relation alignment,
our model correctly predicts the aligned entity pair ($v_2$, $v_5$) and relation pair ($r_2$, $r_5$). After examining the full experimental
data\footnote{A more detailed analysis of our experimental results can be found in Appendix B in the supplementary material.},
we find that the entities with more neighbors, such as $v_2$ and $v_5$ (indicating \emph{Norway}), and the high-frequency relations, such
as $r_2$ and $r_5$ (indicating \emph{country}), are easier to align, since such entities and relations have rich structural information
that can be exploited by a GCN. After jointly learning entity and relation representations, the extra neighboring relation information (e.g., the aligned relations ($r_2$, $r_5$)) enables our model to successfully align $v_{FR}$ and $v_{EN}$. If we keep updating the model to learn better entity and relation representations, our alignment framework can successfully uncover
more entity and relation alignments such as ($v_1$, $v_4$) and ($r_1$, $r_4$). This shows that joint representations can improve both entity and relation alignments.

\section{Conclusions}

This paper presents a novel framework for entity alignment by jointly modeling entities and relations of KGs.
Our approach does not require pre-aligned relations as training data, yet it can simultaneously align entities and relations
of heterogeneous KGs. We achieve this by employing gated GCNs to automatically learn high-quality entity and relation
representations. As a departure from prior work, our approach
constructs joint entity representations that contain both relation information and entity information. We demonstrate that \textit{the
	whole is greater than the sum of its parts}, as the joint representations allow our model to iteratively improve the
learned representations for both entities and relations. Extensive experiments on three real-world datasets show
that our approach delivers better and more robust performance when compared to state-of-the-art methods.

\section*{Acknowledgments}
This work is supported in part by the National Hi-Tech R\&D Program of China (No. 2018YFB1005100), the NSFC Grants (No. 61672057, 61672058, 61872294), and a UK Royal Society International Collaboration Grant (IE161012). For any correspondence, please contact Yansong Feng.

\bibliography{wytemnlp2019}
\bibliographystyle{acl_natbib}

\appendix
\section{Training Procedure of Our Approach}
\label{sec:appendix}

The detailed training procedure of our model is described in Algorithm~\ref{alg:Framwork}.

Our framework takes as input $G_a$ that consists of the two target knowledge graphs (KGs), a set of prior aligned entity pairs. Like a standard neural network training process, we provide a set of tuneable parameters like the number of epochs, $N$. 

As described in Section 4.1 of the main paper, the training of our framework consists of three stages: (1) preliminary entity alignment, (2) approximating relation representations, and (3) joint entity and relation alignment. 

In the first stage, we utilize GCNs to learn entity representation, $\textbf{X}'$, to embed entities of various KGs for preliminary entity alignment.
Next, we use the entity embeddings to
approximate relation representations (i.e., $\bm{r}$ at line 7) to align relations across KGs. When the performance of preliminary entity alignment model has become stable, we enter the third stage (lines 12-23). In this stage, we learn a model to try to incorporate the relation representations into entity embeddings ($\bm{e}_{joint}$ at line 19) to obtain the joint entity representations ($\textbf{X}_{joint}$ at line 20), and continue using GCNs to iteratively integrate neighboring structural information to achieve better entity and relation representations.

\begin{algorithm}[t!]
	\small
	\caption{ Training framework of our model.}
	\label{alg:Framwork}
	\begin{algorithmic}[1]
		\Require
		Graph $G_a$, set of aligned entity pairs $\mathbb{L}$, number of epochs $N$, interval to regenerate negative samples $T$, number of negative samples $\mathcal{K}$.
		\Ensure
		Parameters $\Omega$.
		\State $Mode=Preliminary$, $epoch=0$
		\While{$Model\_has\_not\_stabilized$}
		\State $\textbf{X}'=HighwayGCN(G_a,\Omega)$
		\If{$epoch \bmod T==0$}
		\State $\mathbb{L'}=SampleNegative(\textbf{X}',\mathbb{L},\mathcal{K})$
		\EndIf
		\State $\bm{r} = f(\textbf{H}_r, \textbf{T}_r), \forall r \in R_a$
		\State $L_{pre}=MarginLoss(\textbf{X}',\mathbb{L},\mathbb{L'})$
		\State Back propagate errors and update parameters $\Omega$
		\State $epoch=epoch+1$
		\EndWhile
		\State $Mode=Joint$
		\While{$epoch<N$}
		\State $\textbf{X}'=HighwayGCN(G_a,\Omega)$
		\If{$epoch \bmod T==0$}
		\State $\mathbb{L'}=SampleNegative(\textbf{X}',\mathbb{L},\mathcal{K})$
		\EndIf
		\State $\bm{r} = f(\textbf{H}_r, \textbf{T}_r), \forall r \in R_a$
		\State $\bm{e}_{joint} =  g(\bm{e}, \bm{R}_e), \forall e \in E_a$
		\State $L_{joint}=MarginLoss(\textbf{X}_{joint},\mathbb{L},\mathbb{L'})$
		\State Back propagate errors and update parameters $\Omega$
		\State $epoch=epoch+1$
		\EndWhile
		\State \Return $\Omega$
	\end{algorithmic}
\end{algorithm}


\section{Statistical Information of Alignment}
Table \ref{statics} displays the statistical characteristics of entities and relations which are correctly predicted in both preliminary and joint alignment stage, only in joint alignment stage, and in neither of the stages. From Table \ref{ent}, it can be observed that the more neighbor entities and relations an entity has, the more likely and earlier it will be aligned. So the neighbor information is of great importance to entities, and our model utilizes this information effectively. Also, an entity has higher possibility to be aligned in the joint alignment stage if more entities and relations in its neighborhood have been aligned, for these entities and relations offer precise information and help the embedding of the current entity.

Table \ref{rel} shows the influence of neighbor entities to relation alignment. There is a gap in the frequency of occurrence in triples between relations that are aligned preliminarily or not, because relations with abundant objects are embedded accurately. Moreover, when a relation fails to be aligned in the preliminary stage, it is more likely to be aligned in the joint stage if it appears frequently, as the embedding of relations will improve with the embedding of entities together.

\begin{table}[t!]
	\centering
	\scriptsize
	\begin{subtable}{.5\textwidth}
		\subcaption{Entity Alignment}  
		\label{ent}
		\begin{tabular}{l|c|cccc}
			\Xhline{1pt}
			\multicolumn{2}{c|}{Statics} & Pre$\checkmark$ Joint$\checkmark$ & Pre$\times$ Joint$\checkmark$ & Pre$\times$ Joint$\times$ \\
			\hline
			\multirow{2}{*}{\textbf{\#Nbr Ent.}} & $G_1$ & 7.32 & 6.47 & 6.25\\
			& $G_2$ & 9.52 & 8.51 & 8.33 \\
			\hline
			\multirow{2}{*}{\textbf{\#Nbr Rel.}} & $G_1$ & 3.94 & 3.73 & 3.51 \\
			& $G_2$ & 4.67 & 4.27 & 4.30 \\
			\hline
			\multicolumn{2}{l|}{\textbf{Pre Aligned Ent. (\%)}} & 83 & 79 & 75 \\ 
			\multicolumn{2}{l|}{\textbf{Pre Aligned Rel. (\%)}} & 96 & 95 & 95 \\
			\Xhline{1pt}
		\end{tabular}
	\end{subtable}  
	\par\medskip
	\begin{subtable}{.5\textwidth}
		\subcaption{Relation Alignment}
		\label{rel} 
		\begin{tabular}{l|c|cccc}
			\Xhline{1pt}
			\multicolumn{2}{c|}{Statics} & Pre$\checkmark$ Joint$\checkmark$ & Pre$\times$ Joint$\checkmark$ & Pre$\times$ Joint$\times$ \\
			\hline
			\multirow{2}{*}{\textbf{\quad Freq.\quad}} &\textbf{\ \ \ }$G_1$\textbf{\ \ \ \ } & 85.59 & 17.83 & 16.86\\
			&\textbf{\ \ \ }$G_2$\textbf{\ \ \ \ } & 135.74 & 48.75 & 40.93\\
			\Xhline{1pt}
		\end{tabular}
	\end{subtable}  
	\caption{Statistics of entity and relation alignments. Pre$\checkmark$ Joint$\checkmark$ indicates the set of entities or relations which are predicted correctly in both preliminary and joint alignment, and so on. \#Nbr Ent. and \#Nbr Rel. denote the average number of neighbor entities and relations, respectively. Freq. denotes the average frequency of occurrences of relations in each dataset.}
	\label{statics}
\end{table}

\end{document}